\pdfoutput=1

\documentclass[11pt]{article}

\usepackage[]{ACL2023}

\usepackage{times}
\usepackage{latexsym}

\usepackage[T1]{fontenc}

\usepackage[utf8]{inputenc}

\usepackage{microtype}

\usepackage{inconsolata}

\usepackage{graphicx}
\usepackage{caption}
\usepackage{subcaption}
\usepackage{amsmath}
\usepackage{multirow}
\usepackage{siunitx}

\title{Token-level Fitting Issues of Seq2seq Models}

\author{
    Guangsheng Bao\textsuperscript{\rm 1,2},
    Zhiyang Teng\textsuperscript{\rm 3}, and
    Yue Zhang\thanks{* Corresponding author.} \textsuperscript{\rm ,2,4},
    \\
    \textsuperscript{1} Zhejiang University \\
    \textsuperscript{2} School of Engineering, Westlake University \\
    \textsuperscript{3} Nanyang Technological University \\
    \textsuperscript{4} Institute of Advanced Technology, Westlake Institute for Advanced Study \\
    \textsuperscript{2} \texttt{\{baoguangsheng, zhangyue\}@westlake.edu.cn} \\
    \textsuperscript{3} \texttt{zhiyang.teng@ntu.edu.sg}
}

\begin{document}
\maketitle

\begin{abstract}
Sequence-to-sequence (seq2seq) models have been widely used for natural language processing, computer vision, and other deep learning tasks. We find that seq2seq models trained with early-stopping suffer from issues at the token level. In particular, while some tokens in the vocabulary demonstrate overfitting, others underfit when training is stopped. Experiments show that the phenomena are pervasive in different models, even in fine-tuned large pretrained-models. 
We identify three major factors that influence token-level fitting, which include token frequency, parts-of-speech, and prediction discrepancy. Further, we find that external factors such as language, model size, domain, data scale, and pretraining can also influence the fitting of tokens.

\end{abstract}

\section{Introduction}
Deep learning models tend to overfit on relatively small datasets because of their strong capacity and a massive number of parameters \citep{brownlee2018overfitting, li2019overfitting, rice2020overfitting, bejani2021overfitting}.  Studies suggest regularization and early stopping to control the generalization error caused by overfitting \citep{hastie2009elements, zhang2017understanding, chatterjee2022generalization}. Previous studies mainly analyze the generalization issue on image classification task \citep{arpit17memo, zhang2021understanding}, where the learning target is relatively simple. In contrast, NLP task such as machine translation \citep{zhang2015deep, singh2017machine} is more complex with regard to the learning targets, which involve a sequence of tokens.

\begin{figure*}[t]
  \begin{minipage}{0.66\linewidth}
    \centering
    \begin{subfigure}[b]{0.48\linewidth}
        \centering
        \includegraphics[width=1\linewidth]{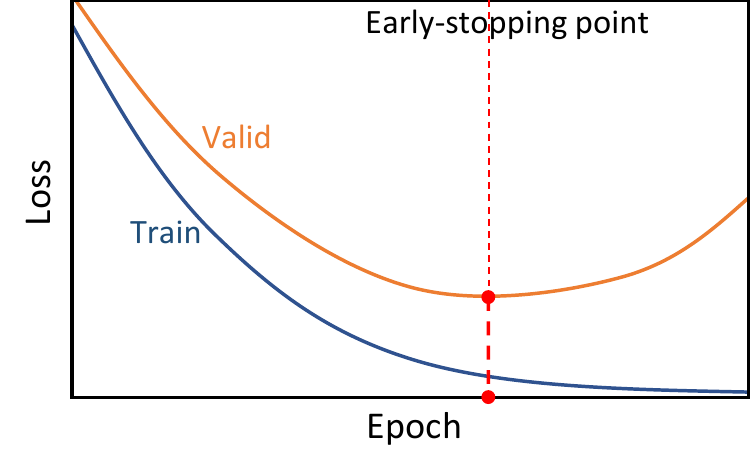}
        \caption{Idealized training and validation loss curves, where the model is selected by early stopping.}
    \end{subfigure}
    \hfill
    \begin{subfigure}[b]{0.48\linewidth}
        \centering
        \includegraphics[width=1\linewidth]{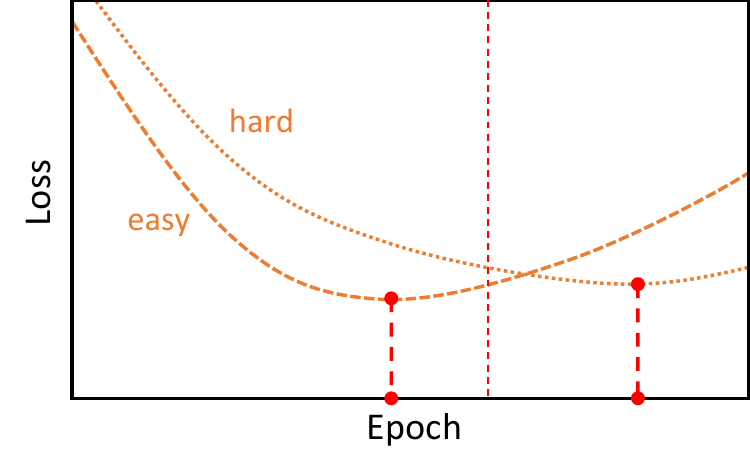}
        \caption{Easy tokens overfit at the early-stopping point, while hard tokens underfit.}
    \end{subfigure}
    \caption{Seq2seq models trained with early stopping may suffer from overfitting or underfitting.}
    \label{fig:overfitting-underfitting}
  \end{minipage}\hfill
  \begin{minipage}{0.317\linewidth}
    \centering
    \includegraphics[width=1\linewidth]{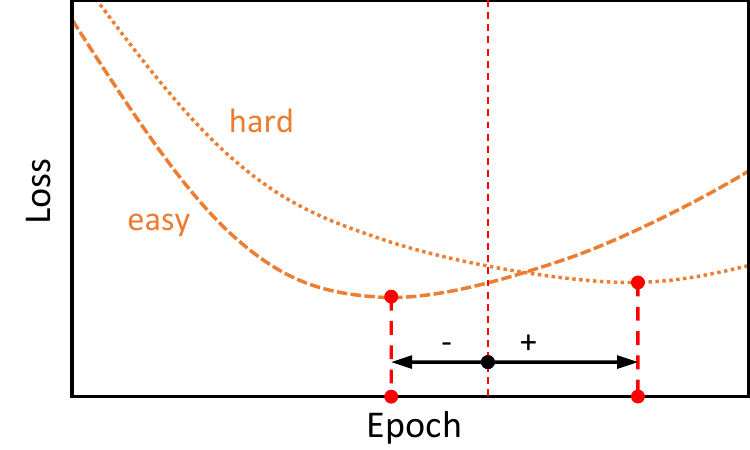}
    \caption{Fitting-offset measures the relative distance of the best fit from the early-stopping point, which is negative for the left side and positive for the right side.}
    \label{fig:fitting-distance}
  \end{minipage}
\end{figure*}

Natural languages exhibit a long-tailed distribution of tokens \citep{powers1998zipf}. The long-tail phenomena have been associated with performance degradation of NLP tasks \citep{gong2018frage, raunak2020long, yu2022rare}, where the rare (low frequency) tokens are ascribed as hard learning targets and popular (high frequency) tokens as easy learning targets. These criteria of easiness of learning targets are intuitive but coarse-grained, which are not associated with the training dynamics. In this paper, we study the easiness of tokens as learning targets from the perspective of overfitting and underfitting. 
Intuitively, the learning on hard tokens will be slower than that on easy tokens, which may result in underfitting on hard tokens and overfitting on easy tokens, as illustrated by Figure \ref{fig:overfitting-underfitting}. We propose two measures to quantify fitting -- \emph{fitting-offset} and \emph{potential-gain}. Fitting-offset measures the offset of the best fit from the early-stopping point, which reflects the degree of overfitting or underfitting. Potential-gain measures the accuracy gap between the early-stopping checkpoint and the best fit, which also estimates the accuracy decrease caused by overfitting or underfitting.

We use machine translation as our test bed, training models on English-German benchmark datasets, including News and Europarl domains. Our extensive experiments reveal novel findings: 1) Both overfitting and underfitting occur in a trained seq2seq model. 2) High-frequency tokens are expected to overfit, but some are found underfitted, and low-frequency tokens are expected to underfit, but some are found overfitted. 3) Large pretrained models reduce underfitting effectively during fine-tuning but are less effective on overfitting. Besides, we propose a direct indicator of easiness -- prediction discrepancy, using the probability difference of predictions made by full context and local context as a criterion to group tokens.

In addition to tokens, sentences have also been considered as learning targets, where curriculum learning methods distinguish sentences as easy or hard \citep{kocmi2017curriculum, platanios2019competence, xu2020dynamic}. For example, the length of a sentence is used as a criterion to identify easy sentences from hard ones, where the easy (short) sentences are first learned and then the hard (long) sentences. Using our metrics, we find that the length of a sentence is not a good indicator of easy or hard sentences from the perspective of overfitting and underfitting.

\section{Related Work}
Previous studies compare the fitting of real data and noise data, demonstrating that real data is easier to learn and has a faster convergence speed than noise data \cite{zhang2017understanding, arpit17memo, chatterjee2022generalization}. The different convergence speeds also provide an explanation of how early stopping prevents the memorization of noise data. However, these works do not compare the fitting among different learning targets inside the real data, neither between different samples nor between different parts of each sample. In this paper, we conduct experiments on a more complex seq2seq task instead of simple classification. We study the fitting of token-level learning targets, demonstrating that both overfitting and underfitting occur when training seq2seq models.

There are few works studying overfitting and underfitting in NLP. \citet{sun2017complex} report that complex structure leads to overfitting in structured prediction. \citet{wolfe2021low} demonstrate that low-frequency names exhibit bias and overfitting in the language model. \citet{varis2021sequence} illustrate that machine translation models generalize poorly on the test set with unseen sentence length. These works discuss overfitting issues on specific conditions, such as complex structure, frequent names, and unseen length. In comparison, we conduct a systematic analysis of the general phenomena of overfitting and underfitting in language. Specifically, we propose quantitative measures, identify major factors, and conduct statistical hypothesis testing on the phenomena.

\section{Experimental Settings}
\subsection{Datasets}
We use the News corpus as a major dataset for our experiments and analysis, and we further use the Europarl corpus for the comparison of different domains and data scales.

\textbf{News}
We use News Commentary v11 for training, newstest2015 for validation, and newstest2016 for testing. The English-German machine translation dataset contains $236,287$ sentence pairs for training, $2,169$ pairs for validation, and $2,999$ pairs for testing.

\textbf{Europarl}
We use English-German Europarl v7, following \citet{maruf2019selective} to split the train, validation, and test sets. The dataset contains $1,666,904$ sentence pairs for training, $3,587$ pairs for validation, and $5,134$ pairs for testing.

We tokenize the sentences using MOSES \cite{koehn2007moses}, followed by truecase and a BPE \cite{sennrich2015neural} with $30,000$ merging operations. We use separate embedding tables for source and target languages in the model.

\subsection{Model Configurations}
We study the overfitting and underfitting issues on three model configurations. We use the base model in section \ref{sec:token-level} and \ref{sec:easy-hard-sent}, the pre-trained model in section \ref{sec:fine-tuning}, and the big model for the model-size evaluation in section \ref{sec:additional-factor}.

\textbf{Base Model} 
We use the standard Transformer base model \cite{vaswani2017attention}, which has 6 layers, 8 heads, 512 output dimensions, and 2048 hidden dimensions. We train the model with a learning rate of $\num{5e-4}$, a dropout of $0.3$, a label smoothing of $0.1$, and an Adam optimizer \cite{kingma2014adam}.

\textbf{Big Model}
Following the standard Transformer big model \cite{vaswani2017attention}, we use 6 layers, 16 heads, 1024 output dimensions, and 4096 hidden dimensions. We train the model with a learning rate of $\num{3e-4}$, a dropout of $0.3$, a label smoothing of $0.1$, and an Adam optimizer.

\textbf{Pretrained Large Model}
We use mBART25 \cite{liu2020multilingual}, which has the similar setting as BART large model \cite{lewis2020bart}, using 12 layers, 16 heads, 1024 output dimensions, and 4096 hidden dimensions. We fine-tune the model with a learning rate of $\num{3e-5}$, a dropout of $0.3$, an attention-dropout of $0.1$, a label smoothing of $0.2$, and an Adam optimizer.

For each experiment, we train $40$ models using random seeds from $1$ to $40$, obtaining $40$ samples for the significance test. During the training of the base or big model, we keep the last $20$ checkpoints for analysis, where the checkpoint of early-stopping is the $10$-th of the $20$ checkpoints. For mBART25, we keep the last $10$ checkpoints, and the early-stopping checkpoint is at the $5$-th of the checkpoints.

\subsection{Evaluation Metric}
\textbf{Measures}
We propose two measures: fitting-offset and potential-gain. 

{\it Fitting-offset} represents how far (i.e., number of epochs) the best fit of a group of tokens diverges from the point of early stopping. In this paper, we use epoch as its unit because we evaluate the model using the validation set at the end of each training epoch. As Figure \ref{fig:fitting-distance} shows, for the easy tokens, the fitting-offset is negative, denoting overfitting, where the best fit is before the early-stopping epoch. For the hard tokens, the fitting-offset is positive, denoting an underfitting, where the best fit is after the early-stopping epoch. Using fitting-offset, we can quantify the degree of overfitting and underfitting.

{\it Potential-gain} represents the potential accuracy increase if we move the best fit to the early-stopping epoch. We calculate the measure by subtracting the accuracy of the early-stopping checkpoint from the accuracy of the best fit. Using this measure, we can quantitatively estimate the potential benefits by fixing the overfitting or underfitting issue.

\textbf{Significance Test}
Since the distribution of fitting-offset is unknown, we use a non-parametric sign-test \cite{dixon1946statistical, hodges1955bivariate} to test our hypothesis. We train the model $N$ times to obtain $N$ observations on the fitting-offset. The hypothesis about the overfitting and underfitting can be expressed by

\begin{equation}
\left\{
\begin{array}{ll}
  H_0: \text{fitting-offset} = 0, & \text{no fitting issue,} \\
  H_1: \text{fitting-offset} \neq 0, & \text{has fitting issue.}
\end{array}
\right.
\label{eq:hypothesis}
\end{equation}

If $H_0$ is true, the $N$ observations are expected to be half positive and half negative. The total number of positive observations $N_+$ follows a binomial distribution, through which we decide the rejection region according to a significance level $\alpha$. 

\textbf{Grouping}
Ideally, we can calculate the two measures on each token to tell which tokens are overfitted and which tokens are underfitted at the early-stopping epoch. However, direct observation of each token is noisy and does not show obvious patterns. We group the tokens and average the valid losses to reduce the noise, through which the pattern emerges, and we obtain stable measures.

\begin{figure}[t]
    \centering
    \includegraphics[width=0.9\linewidth]{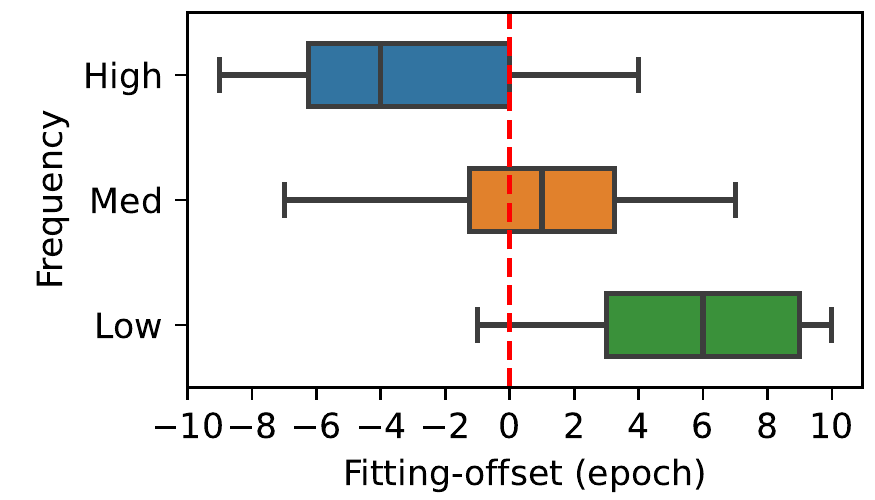}
    \caption{Fitting-offset of tokens grouped by token \emph{frequency}.}
    \label{fig:fitdist-freq}
\end{figure}

\section{Token-level Results}
\label{sec:token-level}

\subsection{Fitting of Rare Tokens in Seq2seq Model Training}
\label{sec:rare-token}
Previous studies suggest that long-tail token distribution affects the performance of NLP tasks \cite{gong2018frage, raunak2020long, yu2022rare}. We hypothesize that the low-frequency tokens underfit during training and conduct verification experiments as follows.

\textbf{Settings}
We experiment on the News dataset, using a Transformer base model \cite{vaswani2017attention}.
We categorize the target tokens into high/medium/low-frequency according to their distribution in the training set, with balanced probability mass on the three buckets.

\begin{figure*}[t]
    \begin{minipage}{0.48\linewidth}
        \centering\small
        \begin{tabular}{lccc}
            \hline
            \bf{Group} & \bf{Parts-of-speech (POS)} \\
            \hline
            Noun & NOUN, PRON, PROPN \\
            Verb & VERB, AUX \\
            Adj & ADJ, ADV \\
            Num & NUM \\
            Func & ADP, CONJ, CCONJ, DET, PART, SCONJ \\
            Symb & PUNCT, SYM \\
            \hline
        \end{tabular}
        \captionof{table}{Groups aggregating parts of speech.}
        \label{tab:parts-of-speech}
    \end{minipage}
    \hfill
    \begin{minipage}{0.48\linewidth}
        \centering
        \includegraphics[width=0.9\linewidth]{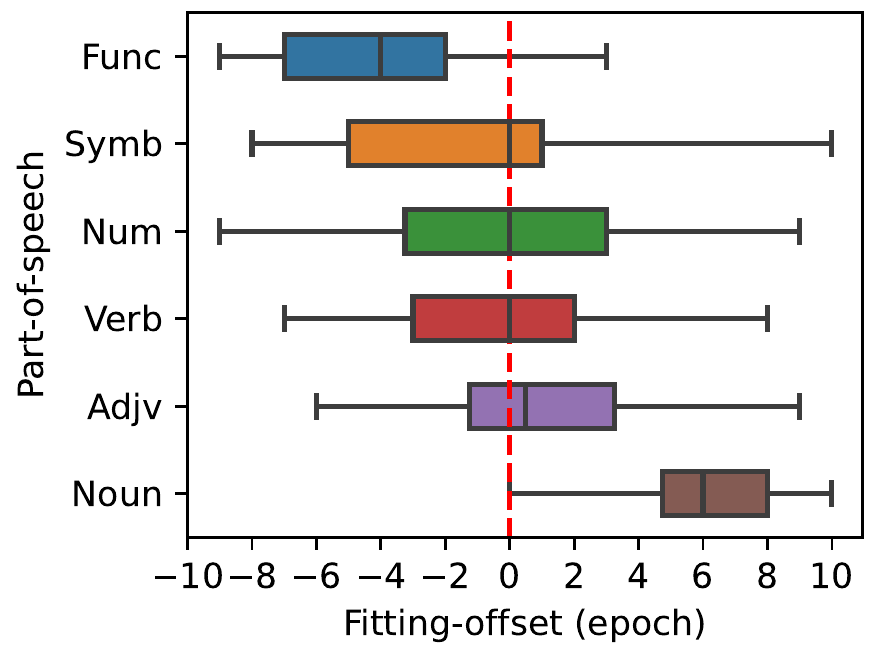}
        \caption{Fitting-offset of tokens grouped by \emph{parts-of-speech}.}
        \label{fig:fitdist-pos}
    \end{minipage}
\end{figure*}

\textbf{Hypothesis Testing}
For each group of high/medium/low-frequency tokens, we measure the fitting-offset using the checkpoints of each model, obtaining $40$ samples for each group. We test our hypothesis on each group using a sign-test (Eq. \ref{eq:hypothesis}). As a result, we obtain a p-value of $\num{1.9e-5}$ for high-frequency and $\num{7.5e-11}$ for low-frequency, which strongly supports the hypothesis that the high/low-frequency tokens either overfit or underfit.

Further as Figure \ref{fig:fitdist-freq} shows, the average fitting-offset for high-frequency tokens is $-3.7$ with a standard deviation of $3.4$. The negative value of the fitting-offset indicates that the high-frequency tokens overfit, where the best fit happens at an average of $3.7$ epochs, before the early-stopping point. The average fitting-offset for low-frequency tokens is $5.8$ with a standard deviation of $3.3$. The positive value of the fitting-offset indicates underfitting, where the best fit happens at $5.8$ epochs, after the early-stopping point on average.
Based on this evidence, we conclude that {\it Both overfitting and underfitting at the token level occur when training seq2seq models.}

\textbf{Analysis}
The significant divergence of fitting-offset between the high/low-frequency tokens suggests that the frequency of tokens has a significant influence on their fitting. We quantify the influence using the potential-gain. In particular, take the low-frequency tokens as an example. The potential-gain is $0.73$, which means that the average accuracy is expected to be increased from $45.61$ to $46.34$ if we move the best fit to the early-stopping epoch.
The potential-gain of the high-frequency tokens is $0.05$, and that of the medium-frequency tokens is $0.23$, which is relatively smaller than that of the low-frequency tokens, suggesting underfitting of the low-frequency tokens is the major issue.

\begin{figure*}[t]
    \centering
    \includegraphics[width=1\linewidth]{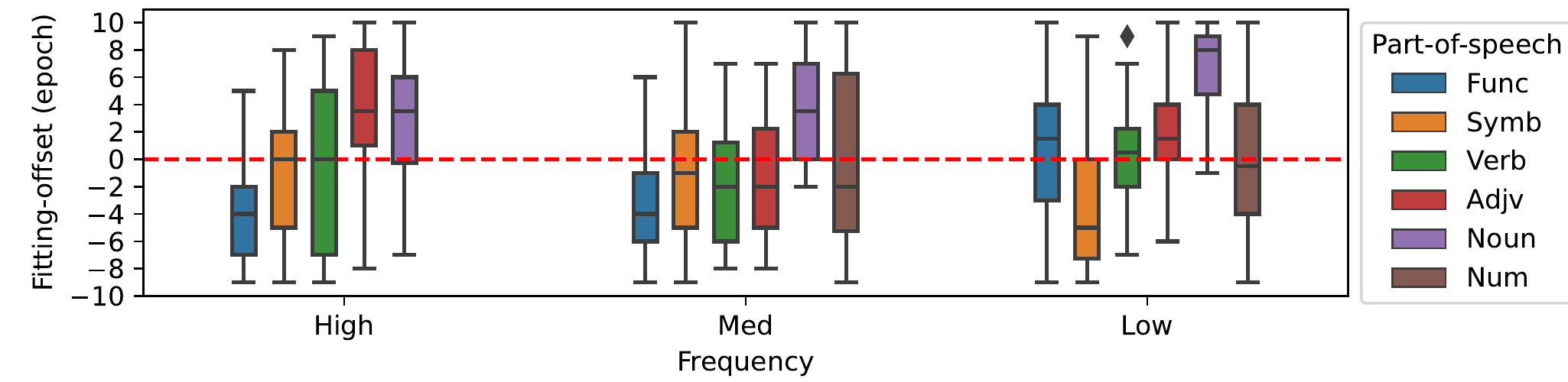}
    \caption{Fitting-offset of tokens grouped by \emph{frequency} and \emph{parts-of-speech}.}
    \label{fig:fitdist-freq-pos}
\end{figure*}

\begin{table*}[t]
    \centering\small
    \begin{tabular}{lcccccc}
        \hline
        \bf{Frequency} & \bf{Function} & \bf{Symbol} & \bf{Number} & \bf{Verb} & \bf{Adj/Adv} & \bf{Noun} \\
        \hline
        High & 56.82 +0.15 & 84.77 \bf{+0.59} & nan +nan & 59.13 \bf{+1.09} & 71.02 \bf{+1.67} & 60.87 \bf{+0.89} \\
        Med & 49.59 +0.16 & 69.63 \bf{+3.41} & 73.47 \bf{+1.50} & 47.64 +0.12 & 44.56 +0.24 & 59.21 \bf{+0.76} \\
        Low & 43.76 \bf{+1.38} & 72.24 \bf{+3.55} & 74.84 \bf{+1.20} & 36.25 \bf{+0.51} & 41.18 +0.43 & 48.03 \bf{+0.88} \\
        \hline
    \end{tabular}
    \caption{Potential-gain for each category grouped by \emph{frequency} and \emph{parts-of-speech}. The column is in a format of ``averaged-accuracy potential-gain'', where the ``+'' in the potential-gain indicates an increase in the accuracy and the ``-'' indicates a decrease in the accuracy. We mark potential-gains bigger than $0.5$ with the bold font to indicate their significance.}
    \label{tab:fitdist-freq-pos}    
\end{table*}

\subsection{Linguistic Factors to Token-level Fitting}
\label{sec:rare-token-underfit}
In section \ref{sec:rare-token}, we find that the high-frequency tokens tend to overfit and the low-frequency tokens tend to underfit in the seq2seq model as a group.
In order to further understand a fine-grained correlation between the frequency and the fitting of a token, we further split the high/low-frequency tokens into smaller groups and conduct experiments on the specific categories. Linguistic factors are considered in the detailed experiments.

\textbf{Parts-of-speech (POS)}
We speculate that parts-of-speech, as an important linguistic feature, may provide a different perspective to study the overfitting and underfitting issues. We group tokens according to their parts-of-speech as listed in Table \ref{tab:parts-of-speech}. 
Specifically, we first obtain POS tagging on each word using spaCy \footnote{https://spacy.io/}. Then we map the POS of words to tokens by labeling all the tokens of a word with the same POS. Last, we group these tokens according to their POS. Take the group Noun as an example. We group tokens with the POS of NOUN, PRON, and PROPN into one category, naming Noun. We aggregate the major parts of speech into six groups according to their functional similarity, as shown in the Table.

As Figure \ref{fig:fitdist-pos} shows, parts-of-speech has a significant influence on the fitting of tokens. The function words are most likely to overfit, which is likely because they are close-set and easier to learn from the linguistic perspective. On the contrary, nouns are most likely to underfit, which can be due to the openness of the set and the challenging context dependencies.

The potential-gain of nouns is $0.69$, increasing the accuracy from $52.38$ to $53.07$. The potential-gains of numbers, symbols, verbs, and adj/adv words are $1.09$, $0.58$, $0.26$, and $0.22$, respectively. Surprisingly, the potential-gain of function words is negligible, even though they obviously overfit. We attribute it to the overall high frequency of function words because sufficient training samples reduce the negative impact of overfitting. It is confirmed by the detailed potential-gains shown in Table \ref{tab:fitdist-freq-pos}, where the function words with low frequency have a much higher potential-gain of $1.38$.

It is worth noting that we use sub-word tokens in the experiments, where we assign the POS of a word to all its sub-words. We also tried words as tokens, which give a similar distribution of fitting offset, and the conclusions hold.

\textbf{Frequency and Parts-of-speech}
We combine frequency and POS to make a detailed analysis of the high/low-frequency tokens.
As Figure \ref{fig:fitdist-freq-pos} shows, frequency and POS work independently. Among the high-frequency tokens, the function words tend to overfit, while adjectives and nouns tend to underfit. Among the low-frequency tokens, the symbols tend to overfit, while the adjectives and nouns tend to underfit. Based on this evidence, we arrive at a counter-intuitive conclusion that
{\it in a seq2seq model, the high-frequency tokens (popular tokens) mostly overfit but can also underfit, and the low-frequency tokens (rare tokens) mostly underfit but can also overfit.}

When we look into the potential-gains, as shown in Table \ref{tab:fitdist-freq-pos}, we see higher potential-gains than in the previous section. The potential-gain of low-frequency function words, symbols, and numbers are $1.38$, $3.55$, and $1.20$, respectively. The potential-gains on med-frequency symbols and numbers are $3.41$ and $1.50$, respectively. Overall the high-frequency tokens have low potential-gains, and the verbs and adjectives have potential-gains of $1.09$ and $1.67$, respectively. These results demonstrate that combining the frequency and linguistic factors reveals stronger overfitting and underfitting, forecasting higher potential-gains in specific categories.

\begin{figure}[t]
    \centering
    \includegraphics[width=0.9\linewidth]{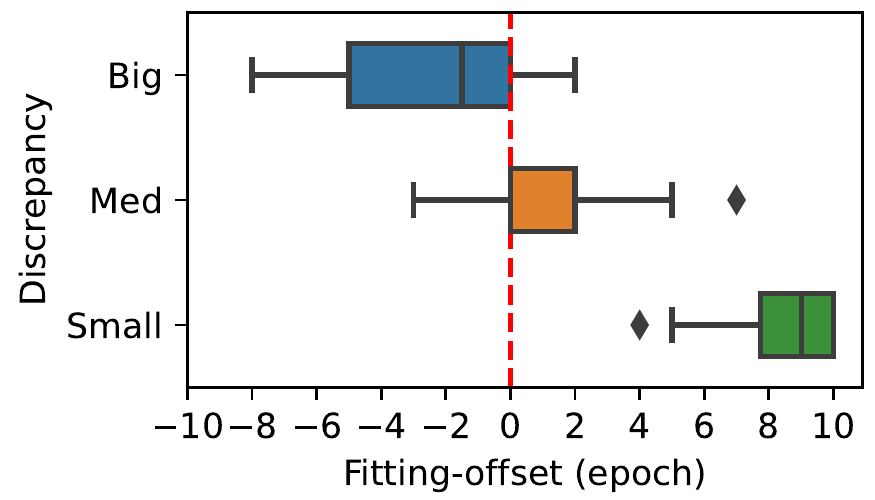}
    \caption{Fitting-offset of tokens grouped by prediction \emph{discrepancy}.}
    \label{fig:fitdist-diff}
\end{figure}

\begin{figure*}[t]
    \begin{minipage}{0.44\linewidth}
        \centering
        \includegraphics[width=1\linewidth]{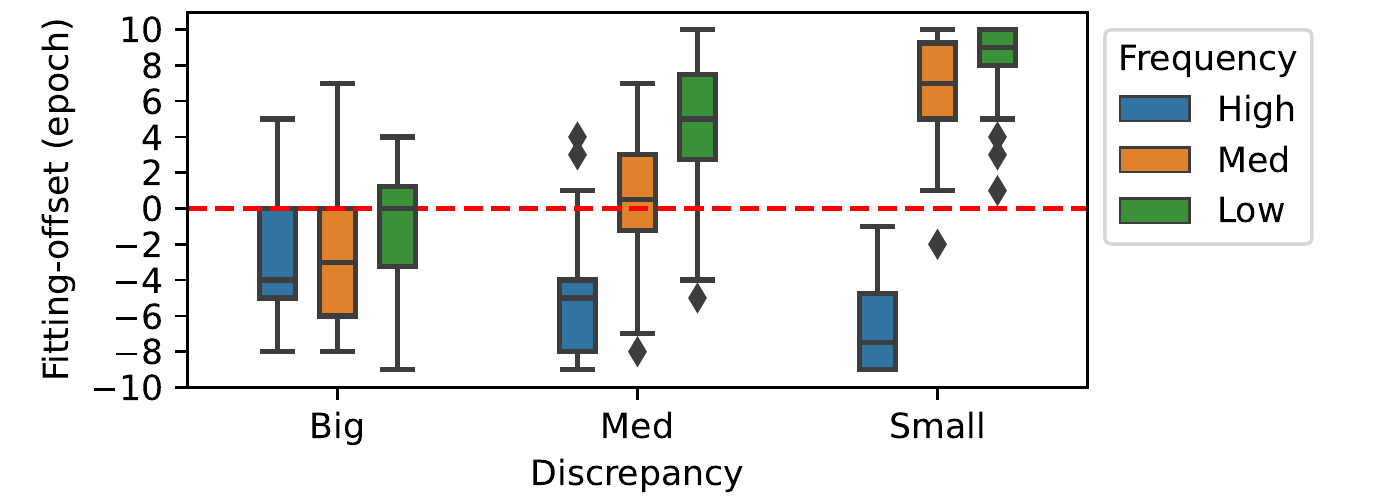}
        \caption{Fitting-offset of tokens grouped by \emph{discrepancy} and \emph{frequency}.}
        \label{fig:fitdist-freq-diff}
    \end{minipage}
    \hfill
    \begin{minipage}{0.54\linewidth}
        \centering\small
        \begin{tabular}{@{}lccc@{}}
            \hline
            \multirow{2}{*}{\bf{Discrepancy}} & \multicolumn{3}{c}{\bf{Frequency}}  \\
             & \bf{High} & \bf{Med} & \bf{Low} \\
            \hline
            Big & 63.91 +0.09 & 26.88 -0.22 & 10.57 +0.11 \\
            Med & 66.96 +0.09 & 47.77 +0.31 & 20.22 \bf{+0.86} \\
            Small & 72.49 +0.09 & 80.95 \bf{+0.62} & 73.19 \bf{+0.85} \\
            \hline
        \end{tabular}
        \captionof{table}{Potential-gain for each category grouped by \emph{discrepancy} and \emph{frequency}.}
        \label{tab:fitdist-freq-diff}
    \end{minipage}
\end{figure*}

\subsection{Prediction Discrepancy as A Factor to Token-level Fitting}
\label{sec:easy-hard-token}
Given that neither frequency nor parts of speech are decisive factors, we propose a new factor -- prediction discrepancy, which measures the degree of dependence on long context for predicting each token in a text sequence. The intuition behind this factor is that some tokens may require more information from the surrounding context to be correctly predicted, while others may be more independent and rely more on their local context. Specifically, consider a sentence, “{\it I was quoted about 12 months .}” the word “{\it months}” could most likely be told according to its local context “{\it 12}”. In comparison, in a German sentence “{\it es war schwer , ihn zu kennen .}”, where the word-by-word translation is “{\it it was hard , him to know .}” the word “{\it kennen}” requires much longer context to predict. 

The prediction discrepancy is calculated as
\begin{equation}
  D_j = |P(Y_j|Y_{<j}, X) - P(Y_j|Y_{j-1}, X)|,
\label{eq:mle-loss}
\end{equation}
where $X$ is the source sequence, and $Y$ is the target. For each token $Y_j$, we predict it using its full context $Y_{<j}$ and its local context $Y_j$. We use the discrepancy between these two predictions to indicate its dependence on long context. We train an altered Transformer model to do the predictions using two decoders, where one decoder uses the full context of a target while another decoder uses the local context. The two decoders share the same token embedding table and encoder. According to the value of discrepancy, we categorize the tokens into three groups, with big, medium, and small discrepancy, respectively. 

\textbf{Results}
As Figure \ref{fig:fitdist-diff} shows, the big-discrepancy tokens have an average fitting-offset of $-2.7$ with a standard deviation of $3.0$. The medium-discrepancy tokens have an average fitting-offset of $1.0$ with a standard deviation of $2.1$. The small-discrepancy tokens have an average fitting-offset of $8.2$ with a standard deviation of $1.8$, showing a trend to exceed the boundary of $10$. 
In comparison with frequency, the bigger range of the average fitting-offsets and the smaller standard deviations suggest that discrepancy is a better indicator than frequency. This indicates that the discrepancy is a good indicator of overfitting and underfitting.

The potential-gain of the small-discrepancy tokens is $0.63$, increasing the average accuracy of the tokens from $75.85$ to $76.48$. In comparison with the potential-gain of $0.75$ for the low-frequency tokens, which increases the average accuracy from $45.61$ to $46.34$, the baseline accuracy of small-discrepancy is much higher, suggesting the effectiveness of discrepancy in discovering fitting issues among high accuracy predictions.

\textbf{Discrepancy and Frequency}
Intuitively, discrepancy and frequency are two independent factors, given that discrepancy relies on context and frequency relies on the token itself. 
As Figure \ref{fig:fitdist-freq-diff} shows, the most significant difference between high-frequency and med/low-frequency tokens is that med/small discrepancy tokens with high frequencies tend to overfit, while the med/small discrepancy tokens with med/low frequencies tend to underfit. 
In addition, as shown in Table \ref{tab:fitdist-freq-diff}, when frequency and discrepancy are combined to predict the overfitting and underfitting, the biggest potential-gain of low-frequency tokens increases from $0.73$ to $0.86$, suggesting that frequency and discrepancy are two independent factors.

\begin{figure*}[t]
    \centering
    \includegraphics[width=1\linewidth]{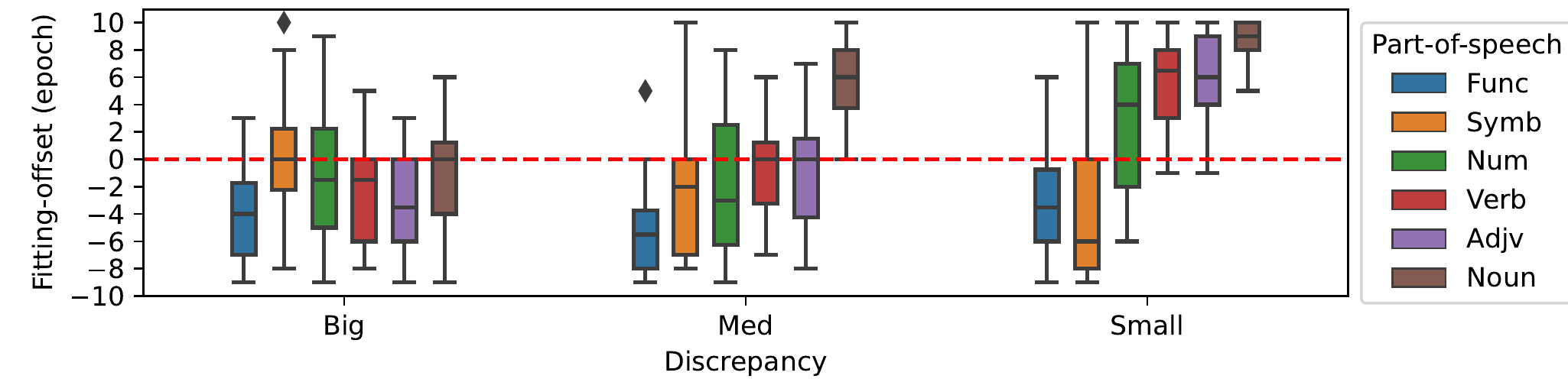}
    \caption{Fitting-offset of tokens grouped by \emph{discrepancy} and \emph{parts-of-speech}.}
    \label{fig:fitdist-diff-pos}
\end{figure*}

\begin{table*}[t]
    \centering\small
    \begin{tabular}{lcccccc}
        \hline
        \bf{Discrepancy} & \bf{Function} & \bf{Symbol} & \bf{Number} & \bf{Verb} & \bf{Adj/Adv} & \bf{Noun} \\
        \hline
        Big & 45.43 -0.07 & 83.38 \bf{+0.51} & 31.58 \bf{+2.72} & 20.31 +0.12 & 13.38 -0.10 & 20.58 -0.06 \\
        Med & 58.33 +0.08 & 78.79 \bf{+0.96} & 36.49 \bf{+2.02} & 39.69 +0.25 & 29.51 +0.14 & 33.30 \bf{+0.93} \\
        Small & 77.06 +0.44 & 77.98 \bf{+1.63} & 89.20 \bf{+1.15} & 69.90 \bf{+0.84} & 71.93 \bf{+0.75} & 77.08 \bf{+0.85} \\
        \hline
    \end{tabular}
    \caption{Potential-gain for each category grouped by \emph{discrepancy} and \emph{parts-of-speech}.}
    \label{tab:fitdist-diff-pos}
\end{table*}

\textbf{Discrepancy and Parts-of-speech (POS)}
As Figure \ref{fig:fitdist-diff-pos} shows, discrepancy and POS also work orthogonally. Overall, tokens with a smaller discrepancy have a larger fitting-offset, which consistently appears on numbers, verbs, adjectives, and nouns. Function words and symbols show a different pattern that the med-discrepancy tokens tend to have smaller fitting-offset than high-discrepancy.

As shown in Table \ref{tab:fitdist-diff-pos}, small-discrepancy tokens have potential gains of $1.63$ and $1.15$ on symbols and numbers, respectively. The potential gain on numbers with big/med-discrepancy are $2.72$ and $2.02$, respectively, suggesting the effectiveness of combining the two factors.

\textbf{Summary}
We have identified three independent factors that affect token-level fitting in seq2seq model training, including frequency, parts-of-speech, and discrepancy. While the former two are internal to the token, the third is external and context-dependent. These indicate that the fitting of tokens results from interestingly complex factors.

\begin{figure}[t]
    \centering
    \includegraphics[width=0.9\linewidth]{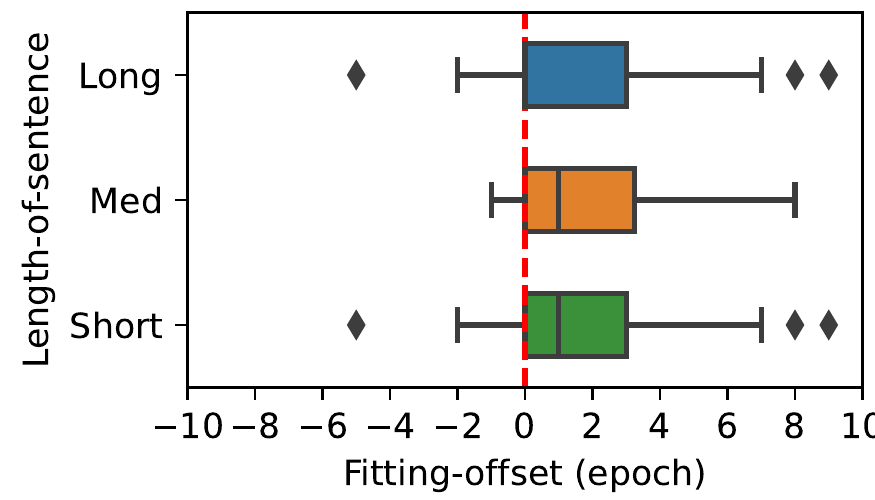}
    \caption{Fitting-offset of tokens grouped by \emph{length-of-sentence}.}
    \label{fig:fitdist-length}
\end{figure}

\section{Easy Sentences vs Hard Sentences}
\label{sec:easy-hard-sent}
Curriculum learning starts with easy sentences and then with hard sentences \cite{kocmi2017curriculum, zhang2018empirical, platanios2019competence, xu2020dynamic, zhou2020uncertainty}, whereas different criteria are used to measure the difficulty of sentences. Among these criteria, the length-of-sentence is the simplest and most popular one, which hypothesizes that short sentences will be easy to learn and long sentences will be hard to learn. We test the hypothesis by evaluating whether short sentences overfit and long sentences underfit in a trained seq2seq model.

\textbf{Length-of-sentence}
We categorize sentences into short/medium/long sentences according to the sentence length that each bucket is allocated with almost the same number of sentences. On the News dataset, the length of short sentences is between $1$ and $18$ tokens, the length of medium sentences between $19$ and $31$, and the length of long sentences between $32$ and $792$.

\textbf{Hypothesis Testing.}
We test our hypothesis using sign-test on News English-German dataset, obtaining a p-value of $\num{3.6e-5}$ for short-sentence, $\num{2.1e-5}$ for medium-sentence, and $\num{2.6e-3}$ for long-sentence, which indicates overfitting or underfitting. The fitting-offset has an average of $1.95$, $2.0$, and $1.38$ for short/medium/long-sentences, respectively. The positive fitting-offsets suggest that they overfit in the trained models. However, as Figure \ref{fig:fitdist-length} shows, the degree of overfitting and underfitting is less than that of frequency (Figure \ref{fig:fitdist-freq}) and discrepancy (Figure \ref{fig:fitdist-diff}).

\textbf{Summary}
The above experiments suggest that although the length-of-sentence can differentiate easy sentences from hard sentences, its effectiveness may not be as significant as other factors such as frequency, discrepancy, and parts-of-speech. More surprisingly, short sentences are more likely to underfit than long sentences, which is also confirmed by experiments on pretraining settings in section \ref{sec:fine-tuning}, suggesting that we could not simply judge the short-sentences as easy and long-sentences as hard.

\begin{figure*}[t]
    \centering 
    \begin{subfigure}[b]{0.66\linewidth}
    \begin{subfigure}[b]{0.49\linewidth}
        \centering
        \includegraphics[width=1\linewidth]{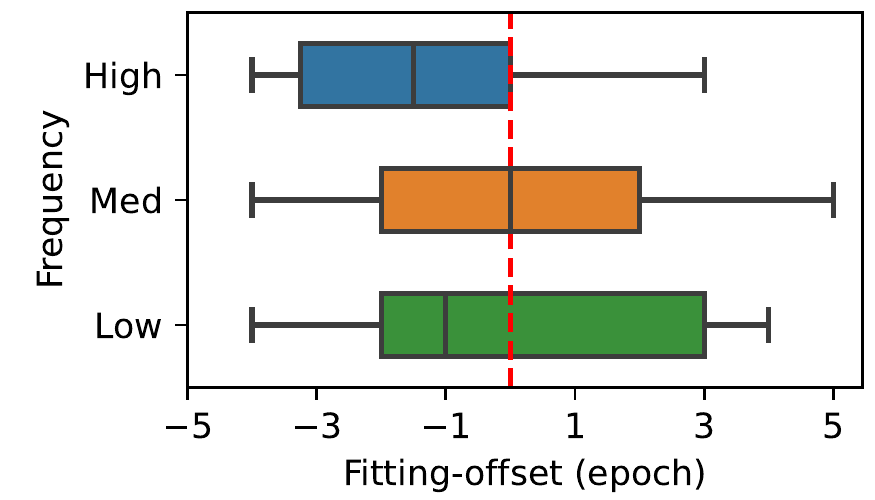}
        \caption{Fitting-offset of tokens grouped by \emph{frequency}.}
        \label{fig:fitdist-freq-mbart}
    \end{subfigure}
    \hfill
    \begin{subfigure}[b]{0.49\linewidth}
        \centering
        \includegraphics[width=1\linewidth]{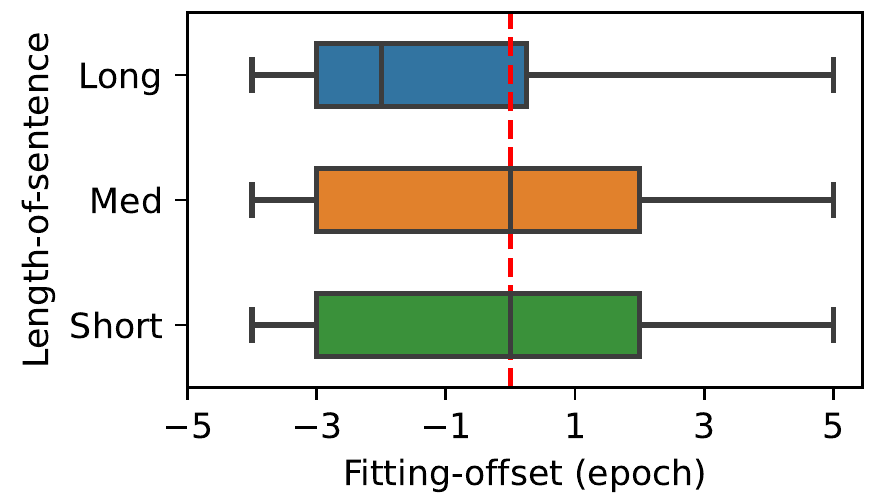}
        \caption{Fitting-offset of tokens grouped by \emph{length-of-sentence}.}
        \label{fig:fitdist-length-mbart}
    \end{subfigure}
    \end{subfigure}
    \hfill
    \begin{subfigure}[b]{0.33\linewidth}
        \centering
        \includegraphics[width=1\linewidth]{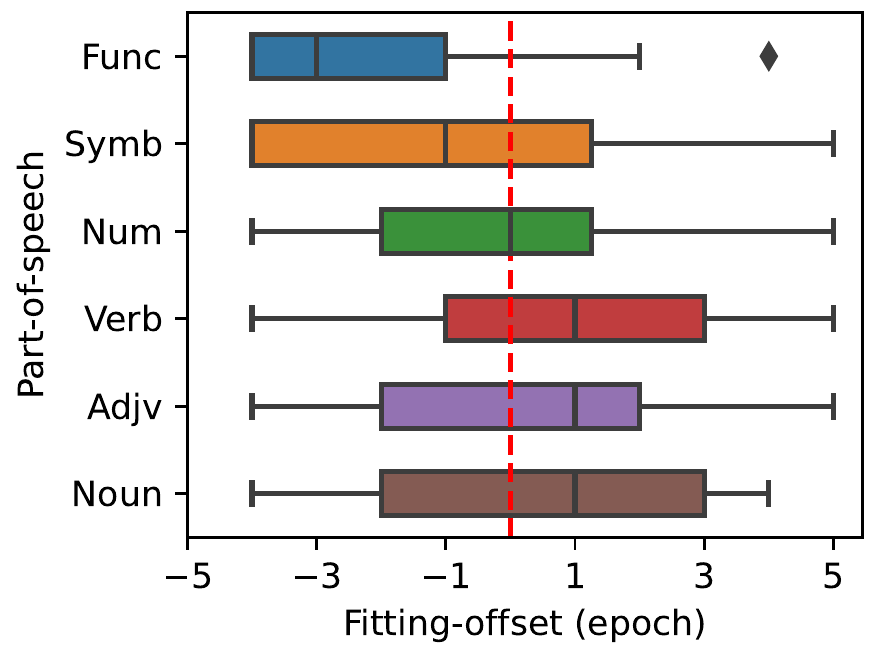}
        \caption{Fitting-offset of tokens grouped by \emph{parts-of-speech}.}
        \label{fig:fitdist-pos-mbart}
    \end{subfigure}
    \caption{The distribution of fitting-offset on \emph{pretraining} setting.}
    \label{fig:fitdist-mbart}
\end{figure*}

\section{Fine-tuning of Pretrained Language Models}
\label{sec:fine-tuning}
Fine-tuning on a large pretrained model has become the dominant setting for NLP tasks \cite{kenton2019bert, lewis2020bart, brown2020language, liu2020multilingual}. We investigate the overfitting and underfitting issues, particularly in the pretraining setting.

\textbf{Hypothesis Testing}
We first test whether the overfitting and underfitting issues exist under the pretraining setting. We experiment by fine-tuning mBART25 on the News English-German dataset. 

First, we evaluate frequency as an indicator of overfitting and underfitting. As Figure \ref{fig:fitdist-freq-mbart} shows, we obtain a p-value of $\num{1.2e-3}$ and an average fitting-offset of $-1.48$ on high-frequency tokens, suggesting the tendency of overfitting on high-frequency tokens. Results on medium/low-frequency tokens do not show significance, although the average fitting-offsets of $0.08$ on medium-frequency tokens suggest slight underfitting.

Next, we consider length-of-sentence. As Figure \ref{fig:fitdist-length-mbart} shows, the fine-tuning tends to overfit for long-sentence and we obtain a p-value of $\num{1.1e-2}$ and an average fitting-offset of $-1.2$ on long-sentences. It suggests obvious overfitting of long sentences, which is counter-intuitive because it is widely assumed that long-sentence is harder to learn than short-sentence.

Last, we examine parts of speech. As Figure \ref{fig:fitdist-pos-mbart} shows, fine-tuning on function words tends to overfit and that on nouns underfit, which is consistent with the result on non-pretraining setting (Figure \ref{fig:fitdist-pos}). A difference is that the fine-tuning of verbs shows underfitting, which is not significantly observed in the non-pretraining settings. We obtain a p-value of $\num{5.8e-2}$ and an average fitting-offset of $0.8$ on verbs, suggesting an observable underfitting on verbs. The issue on function words is more significant than on verbs, on which we obtain a p-value of $\num{4.3e-6}$ and an average fitting-offset of $-2.3$, suggesting significant overfitting on function words. 

The experiments above confirm that the overfitting and underfitting issues exist in the pretraining setting, although it is not as significant as that in the non-pretraining settings. In addition, it shows that overfitting is the major issue in comparison with underfitting. We attribute it to the effectiveness of large pretraining to prevent underfitting.

It is worth noting that the vocabulary used by the pre-trained model is different from other non-pretraining settings, which may affect the statistical figures but the conclusions hold. 

\section{Additional Factors to the Fitting Issue}
\label{sec:additional-factor}
Most of the figures and tables for this section are placed in Appendix \ref{app:additional-factors}.

\textbf{The Language}
As a comparison of languages, we study the issues on the News dataset but with a reversed direction of languages, translating German to English instead of English to German. As Figure \ref{fig:fitdist-de2en} shows, our observations on target English tokens are consistent with previous observations on target German tokens.
First, the high-frequency tokens tend to overfit, while the low-frequency tokens tend to underfit.
Second, the big-discrepancy tokens tend to overfit, while the small-discrepancy tokens tend to underfit.
Third, the function words tend to overfit, while the nouns tend to underfit.
Last, we also obtain bigger potential-gains by combining the factors and the most significant potential-gains happen on the consistent categories, such as low-frequency symbols, high-frequency adj/advs, and big/med-discrepancy numbers, as the bold numbers in Table \ref{tab:fitdist-freq-diff-de2en}, \ref{tab:fitdist-freq-pos-de2en}, and \ref{tab:fitdist-diff-pos-de2en} show.

\textbf{The Model Size}
To evaluate the influence of model size, we conduct experiments with a big model on the News English-German dataset.
As Figure \ref{fig:fitdist-en2de-big} shows, the distribution of fitting-offset on each category is very close to that of the base model but with a smaller range. We attribute it to the faster convergence of bigger model \cite{li2020train}.
One significant difference between the big model and the base model is that the fitting-offset of symbols moves toward the negative region, suggesting overfitting for symbols in the big model. We attribute it to the stronger memorization ability of the bigger model.

\textbf{The Domain}
Previous experiments are done on the News dataset. To justify that the phenomena are not domain specific, we conduct the same experiments on the Europarl English-German dataset, which is from the Europarl domain. We randomly sample $250,000$ sentence pairs from the Europarl training set for a fair comparison with the News dataset, which contains $236,287$ samples.

As Figure \ref{fig:fitdist-en2de-europarl25w} shows, the observations described in ``The Language'' block hold but with slight differences in the distribution of fitting-offset of tokens grouped by parts-of-speech, in comparison with Figure \ref{fig:fitdist-pos} evaluated on News dataset. The major difference happens in verbs, adjectives, and nouns, reflecting a different distribution of topics of Europarl in comparison with News.

\textbf{The Data Scale}
Intuitively, overfitting and underfitting should be more severe on small datasets. For comparison, we test on a bigger dataset from Europarl. We sample $500,000$ sentence pairs from the Europarl training set in comparison with the model trained using $250,000$ samples.

As Figure \ref{fig:fitdist-en2de-europarl50w} shows, the observations described in ``The Language'' block still hold but the range of the distribution increases. Looking into the potential-gains, we see that they decrease by about $1/4$ compared to that of the experiments with $250,000$ samples. The results suggest that the fitting-offset is more challenging to measure, and the potential-gain decreases when the model is trained on a larger dataset, which is expected due to the larger dataset reducing overfitting and underfitting.

\section{Discussion on Possible Solutions}
Our observation and insights are based on \emph{general language phenomena}: long-tail distribution, uneven context dependencies, and different parts of speech, which are general among different NLP tasks, and may deserve further study on possible solutions.

Intuitively, the model training or fine-tuning process could be improved by reducing learning on overfitting tokens and increasing learning on underfitting tokens. One example is the previous work on abstractive summarization \cite{shi2018normalized}, which uses focal loss to give high-freq tokens smaller weights and low-freq tokens larger weights, achieved 1.5 points improvement on ROUGE-2 on the abstractive summarization task. Such \emph{dynamic token-specific weights} can partially address the underfitting and overfitting from the perspective of the frequency factor. However, the issue is far from being solved, given that a combination of the frequency and discrepancy factors (Figure \ref{fig:fitdist-freq-diff}) reveals more severe fitting issues than the frequency only (Figure  \ref{fig:fitdist-freq}). We could expect more gains by designing more appropriate methods to consider the combined factors.

\section{Conclusion}
We study overfitting and underfitting issues of learning targets in the context of neural machine translation. Our experiments demonstrate that overall rare tokens tend to underfit and frequent tokens overfit. We explored detailed factors related to the overfitting and underfitting issues and identified three major influencing factors, which include frequency, parts-of-speech, and discrepancy. This shows that fitting is the result of a complex interaction between multiple factors. Further experiments demonstrate that the issues exist as a general problem for both non-pretraining and pretraining settings. Future work includes the investigation of strategies to alleviate the overfitting and underfitting issues. 

\section*{Limitations}
This paper hypothesizes that seq2seq models have token-level overfitting and underfitting issues, and provides direct evidence to support the hypothesis in various settings, raising a valuable problem for NLP modeling. However, this paper does not provide a solution to the problem due to the theoretical and practical challenges of measuring the convergence speed of each token. We leave the exploration of this topic to future work.

\section*{Acknowledgements}
We would like to thank the anonymous reviewers for their valuable feedback. This work is funded by the China Strategic Scientific and Technological Innovation Cooperation Project (grant No. SQ2022YFE020038) and the National Natural Science Foundation of China (grant NSFC No. 62161160339). Zhiyang Teng is partially supported by CAAI-Huawei MindSpore Open Fund (CAAIXSJLJJ-2021-046A).

\bibliography{custom}

\begin{thebibliography}{36}
\expandafter\ifx\csname natexlab\endcsname\relax\def\natexlab#1{#1}\fi

\bibitem[{Arpit et~al.(2017)Arpit, Jastrz{\k{e}}bski, Ballas, Krueger, Bengio,
  Kanwal, Maharaj, Fischer, Courville, Bengio, and
  Lacoste-Julien}]{arpit17memo}
Devansh Arpit, Stanis{\l}aw Jastrz{\k{e}}bski, Nicolas Ballas, David Krueger,
  Emmanuel Bengio, Maxinder~S. Kanwal, Tegan Maharaj, Asja Fischer, Aaron
  Courville, Yoshua Bengio, and Simon Lacoste-Julien. 2017.
\newblock A closer look at memorization in deep networks.
\newblock In \emph{Proceedings of the 34th International Conference on Machine
  Learning}, volume~70 of \emph{Proceedings of Machine Learning Research},
  pages 233--242. PMLR.

\bibitem[{Bejani and Ghatee(2021)}]{bejani2021overfitting}
Mohammad~Mahdi Bejani and Mehdi Ghatee. 2021.
\newblock A systematic review on overfitting control in shallow and deep neural
  networks.
\newblock \emph{Artificial Intelligence Review}, 54(8):6391--6438.

\bibitem[{Brown et~al.(2020)Brown, Mann, Ryder, Subbiah, Kaplan, Dhariwal,
  Neelakantan, Shyam, Sastry, Askell et~al.}]{brown2020language}
Tom Brown, Benjamin Mann, Nick Ryder, Melanie Subbiah, Jared~D Kaplan, Prafulla
  Dhariwal, Arvind Neelakantan, Pranav Shyam, Girish Sastry, Amanda Askell,
  et~al. 2020.
\newblock Language models are few-shot learners.
\newblock \emph{Advances in neural information processing systems},
  33:1877--1901.

\bibitem[{Brownlee(2018)}]{brownlee2018overfitting}
Jason Brownlee. 2018.
\newblock \emph{Better deep learning: train faster, reduce overfitting, and
  make better predictions}.
\newblock Machine Learning Mastery.

\bibitem[{Chatterjee and Zielinski(2022)}]{chatterjee2022generalization}
Satrajit Chatterjee and Piotr Zielinski. 2022.
\newblock On the generalization mystery in deep learning.
\newblock \emph{arXiv preprint arXiv:2203.10036}.

\bibitem[{Dixon and Mood(1946)}]{dixon1946statistical}
Wilfrid~J Dixon and Alexander~M Mood. 1946.
\newblock The statistical sign test.
\newblock \emph{Journal of the American Statistical Association},
  41(236):557--566.

\bibitem[{Gong et~al.(2018)Gong, He, Tan, Qin, Wang, and Liu}]{gong2018frage}
Chengyue Gong, Di~He, Xu~Tan, Tao Qin, Liwei Wang, and Tie-Yan Liu. 2018.
\newblock Frage: Frequency-agnostic word representation.
\newblock \emph{Advances in neural information processing systems}, 31.

\bibitem[{Hastie et~al.(2009)Hastie, Tibshirani, Friedman, and
  Friedman}]{hastie2009elements}
Trevor Hastie, Robert Tibshirani, Jerome~H Friedman, and Jerome~H Friedman.
  2009.
\newblock \emph{The elements of statistical learning: data mining, inference,
  and prediction}, volume~2.
\newblock Springer.

\bibitem[{Hodges(1955)}]{hodges1955bivariate}
Joseph~L Hodges. 1955.
\newblock A bivariate sign test.
\newblock \emph{The Annals of Mathematical Statistics}, 26(3):523--527.

\bibitem[{Kenton and Toutanova(2019)}]{kenton2019bert}
Jacob Devlin Ming-Wei~Chang Kenton and Lee~Kristina Toutanova. 2019.
\newblock Bert: Pre-training of deep bidirectional transformers for language
  understanding.
\newblock In \emph{Proceedings of NAACL-HLT}, pages 4171--4186.

\bibitem[{Kingma and Ba(2014)}]{kingma2014adam}
Diederik~P Kingma and Jimmy Ba. 2014.
\newblock Adam: A method for stochastic optimization.
\newblock \emph{arXiv preprint arXiv:1412.6980}.

\bibitem[{Kocmi and Bojar(2017)}]{kocmi2017curriculum}
Tom Kocmi and Ond{\v{r}}ej Bojar. 2017.
\newblock Curriculum learning and minibatch bucketing in neural machine
  translation.
\newblock In \emph{Proceedings of the International Conference Recent Advances
  in Natural Language Processing, RANLP 2017}, pages 379--386.

\bibitem[{Koehn et~al.(2007)Koehn, Hoang, Birch, Callison-Burch, Federico,
  Bertoldi, Cowan, Shen, Moran, Zens et~al.}]{koehn2007moses}
Philipp Koehn, Hieu Hoang, Alexandra Birch, Chris Callison-Burch, Marcello
  Federico, Nicola Bertoldi, Brooke Cowan, Wade Shen, Christine Moran, Richard
  Zens, et~al. 2007.
\newblock Moses: Open source toolkit for statistical machine translation.
\newblock In \emph{Proceedings of the 45th annual meeting of the association
  for computational linguistics companion volume proceedings of the demo and
  poster sessions}, pages 177--180.

\bibitem[{Lewis et~al.(2020)Lewis, Liu, Goyal, Ghazvininejad, Mohamed, Levy,
  Stoyanov, and Zettlemoyer}]{lewis2020bart}
Mike Lewis, Yinhan Liu, Naman Goyal, Marjan Ghazvininejad, Abdelrahman Mohamed,
  Omer Levy, Veselin Stoyanov, and Luke Zettlemoyer. 2020.
\newblock Bart: Denoising sequence-to-sequence pre-training for natural
  language generation, translation, and comprehension.
\newblock In \emph{Proceedings of the 58th Annual Meeting of the Association
  for Computational Linguistics}, pages 7871--7880.

\bibitem[{Li et~al.(2019)Li, Li, Guan, Liang, Lai, and Luo}]{li2019overfitting}
Haidong Li, Jiongcheng Li, Xiaoming Guan, Binghao Liang, Yuting Lai, and
  Xinglong Luo. 2019.
\newblock Research on overfitting of deep learning.
\newblock In \emph{2019 15th International Conference on Computational
  Intelligence and Security (CIS)}, pages 78--81. IEEE.

\bibitem[{Li et~al.(2020)Li, Wallace, Shen, Lin, Keutzer, Klein, and
  Gonzalez}]{li2020train}
Zhuohan Li, Eric Wallace, Sheng Shen, Kevin Lin, Kurt Keutzer, Dan Klein, and
  Joey Gonzalez. 2020.
\newblock Train big, then compress: Rethinking model size for efficient
  training and inference of transformers.
\newblock In \emph{International Conference on Machine Learning}, pages
  5958--5968. PMLR.

\bibitem[{Liu et~al.(2020)Liu, Gu, Goyal, Li, Edunov, Ghazvininejad, Lewis, and
  Zettlemoyer}]{liu2020multilingual}
Yinhan Liu, Jiatao Gu, Naman Goyal, Xian Li, Sergey Edunov, Marjan
  Ghazvininejad, Mike Lewis, and Luke Zettlemoyer. 2020.
\newblock Multilingual denoising pre-training for neural machine translation.
\newblock \emph{Transactions of the Association for Computational Linguistics},
  8:726--742.

\bibitem[{Maruf et~al.(2019)Maruf, Martins, and Haffari}]{maruf2019selective}
Sameen Maruf, Andr{\'e}~FT Martins, and Gholamreza Haffari. 2019.
\newblock Selective attention for context-aware neural machine translation.
\newblock In \emph{Proceedings of the 2019 Conference of the North American
  Chapter of the Association for Computational Linguistics: Human Language
  Technologies, Volume 1 (Long and Short Papers)}, pages 3092--3102.

\bibitem[{Platanios et~al.(2019)Platanios, Stretcu, Neubig, Poczos, and
  Mitchell}]{platanios2019competence}
Emmanouil~Antonios Platanios, Otilia Stretcu, Graham Neubig, Barnabas Poczos,
  and Tom~M Mitchell. 2019.
\newblock Competence-based curriculum learning for neural machine translation.
\newblock In \emph{Proceedings of NAACL-HLT}, pages 1162--1172.

\bibitem[{Powers(1998)}]{powers1998zipf}
David~MW Powers. 1998.
\newblock Applications and explanations of zipf’s law.
\newblock In \emph{New methods in language processing and computational natural
  language learning}.

\bibitem[{Raunak et~al.(2020)Raunak, Dalmia, Gupta, and Metze}]{raunak2020long}
Vikas Raunak, Siddharth Dalmia, Vivek Gupta, and Florian Metze. 2020.
\newblock On long-tailed phenomena in neural machine translation.
\newblock In \emph{Findings of the Association for Computational Linguistics:
  EMNLP 2020}, pages 3088--3095.

\bibitem[{Rice et~al.(2020)Rice, Wong, and Kolter}]{rice2020overfitting}
Leslie Rice, Eric Wong, and Zico Kolter. 2020.
\newblock Overfitting in adversarially robust deep learning.
\newblock In \emph{International Conference on Machine Learning}, pages
  8093--8104. PMLR.

\bibitem[{Sennrich et~al.(2015)Sennrich, Haddow, and
  Birch}]{sennrich2015neural}
Rico Sennrich, Barry Haddow, and Alexandra Birch. 2015.
\newblock Neural machine translation of rare words with subword units.
\newblock \emph{arXiv preprint arXiv:1508.07909}.

\bibitem[{Shi et~al.(2018)Shi, Meng, Wang, Lin, and Li}]{shi2018normalized}
Yunsheng Shi, Jun Meng, Jian Wang, Hongfei Lin, and Yumeng Li. 2018.
\newblock A normalized encoder-decoder model for abstractive summarization
  using focal loss.
\newblock In \emph{Natural Language Processing and Chinese Computing: 7th CCF
  International Conference, NLPCC 2018, Hohhot, China, August 26--30, 2018,
  Proceedings, Part II 7}, pages 383--392. Springer.

\bibitem[{Singh et~al.(2017)Singh, Kumar, Darbari, Singh, Rastogi, and
  Jain}]{singh2017machine}
Shashi~Pal Singh, Ajai Kumar, Hemant Darbari, Lenali Singh, Anshika Rastogi,
  and Shikha Jain. 2017.
\newblock Machine translation using deep learning: An overview.
\newblock In \emph{2017 international conference on computer, communications
  and electronics (comptelix)}, pages 162--167. IEEE.

\bibitem[{Sun et~al.(2017)Sun, Sun, Ma, Ren, Zhang, Li, and
  Wang}]{sun2017complex}
Xu~Sun, Weiwei Sun, Shuming Ma, Xuancheng Ren, Yi~Zhang, Wenjie Li, and Houfeng
  Wang. 2017.
\newblock Complex structure leads to overfitting: A structure regularization
  decoding method for natural language processing.
\newblock \emph{arXiv preprint arXiv:1711.10331}.

\bibitem[{Varis and Bojar(2021)}]{varis2021sequence}
Dusan Varis and Ond{\v{r}}ej Bojar. 2021.
\newblock Sequence length is a domain: Length-based overfitting in transformer
  models.
\newblock In \emph{Proceedings of the 2021 Conference on Empirical Methods in
  Natural Language Processing}, pages 8246--8257.

\bibitem[{Vaswani et~al.(2017)Vaswani, Shazeer, Parmar, Uszkoreit, Jones,
  Gomez, Kaiser, and Polosukhin}]{vaswani2017attention}
Ashish Vaswani, Noam Shazeer, Niki Parmar, Jakob Uszkoreit, Llion Jones,
  Aidan~N Gomez, {\L}ukasz Kaiser, and Illia Polosukhin. 2017.
\newblock Attention is all you need.
\newblock \emph{Advances in neural information processing systems}, 30.

\bibitem[{Wolfe and Caliskan(2021)}]{wolfe2021low}
Robert Wolfe and Aylin Caliskan. 2021.
\newblock Low frequency names exhibit bias and overfitting in contextualizing
  language models.
\newblock In \emph{Proceedings of the 2021 Conference on Empirical Methods in
  Natural Language Processing}, pages 518--532.

\bibitem[{Xu et~al.(2020)Xu, Hu, Jiang, Feng, Wang, Huang, Ju, Xiao, and
  Zhu}]{xu2020dynamic}
Chen Xu, Bojie Hu, Yufan Jiang, Kai Feng, Zeyang Wang, Shen Huang, Qi~Ju, Tong
  Xiao, and Jingbo Zhu. 2020.
\newblock Dynamic curriculum learning for low-resource neural machine
  translation.
\newblock In \emph{Proceedings of the 28th International Conference on
  Computational Linguistics}, pages 3977--3989.

\bibitem[{Yu et~al.(2022)Yu, Song, Kim, Lee, Ryu, and Yoon}]{yu2022rare}
Sangwon Yu, Jongyoon Song, Heeseung Kim, Seongmin Lee, Woo-Jong Ryu, and
  Sungroh Yoon. 2022.
\newblock Rare tokens degenerate all tokens: Improving neural text generation
  via adaptive gradient gating for rare token embeddings.
\newblock In \emph{Proceedings of the 60th Annual Meeting of the Association
  for Computational Linguistics (Volume 1: Long Papers)}, pages 29--45.

\bibitem[{Zhang et~al.(2017)Zhang, Bengio, Hardt, Recht, and
  Vinyals}]{zhang2017understanding}
Chiyuan Zhang, Samy Bengio, Moritz Hardt, Benjamin Recht, and Oriol Vinyals.
  2017.
\newblock Understanding deep learning requires rethinking generalization.
\newblock In \emph{International Conference on Learning Representations}.

\bibitem[{Zhang et~al.(2021)Zhang, Bengio, Hardt, Recht, and
  Vinyals}]{zhang2021understanding}
Chiyuan Zhang, Samy Bengio, Moritz Hardt, Benjamin Recht, and Oriol Vinyals.
  2021.
\newblock Understanding deep learning (still) requires rethinking
  generalization.
\newblock \emph{Communications of the ACM}, 64(3):107--115.

\bibitem[{Zhang et~al.(2015)Zhang, Zong et~al.}]{zhang2015deep}
Jiajun Zhang, Chengqing Zong, et~al. 2015.
\newblock Deep neural networks in machine translation: An overview.
\newblock \emph{IEEE Intell. Syst.}, 30(5):16--25.

\bibitem[{Zhang et~al.(2018)Zhang, Kumar, Khayrallah, Murray, Gwinnup,
  Martindale, McNamee, Duh, and Carpuat}]{zhang2018empirical}
Xuan Zhang, Gaurav Kumar, Huda Khayrallah, Kenton Murray, Jeremy Gwinnup,
  Marianna~J Martindale, Paul McNamee, Kevin Duh, and Marine Carpuat. 2018.
\newblock An empirical exploration of curriculum learning for neural machine
  translation.
\newblock \emph{arXiv preprint arXiv:1811.00739}.

\bibitem[{Zhou et~al.(2020)Zhou, Yang, Wong, Wan, and
  Chao}]{zhou2020uncertainty}
Yikai Zhou, Baosong Yang, Derek~F Wong, Yu~Wan, and Lidia~S Chao. 2020.
\newblock Uncertainty-aware curriculum learning for neural machine translation.
\newblock In \emph{Proceedings of the 58th Annual Meeting of the Association
  for Computational Linguistics}, pages 6934--6944.

\end{thebibliography}
\bibliographystyle{acl_natbib}

\appendix
\section{Appendix}
\label{app:additional-factors}

\begin{figure*}[t]
    \centering    
    \begin{subfigure}[b]{0.325\linewidth}
        \centering
        \includegraphics[width=1\linewidth]{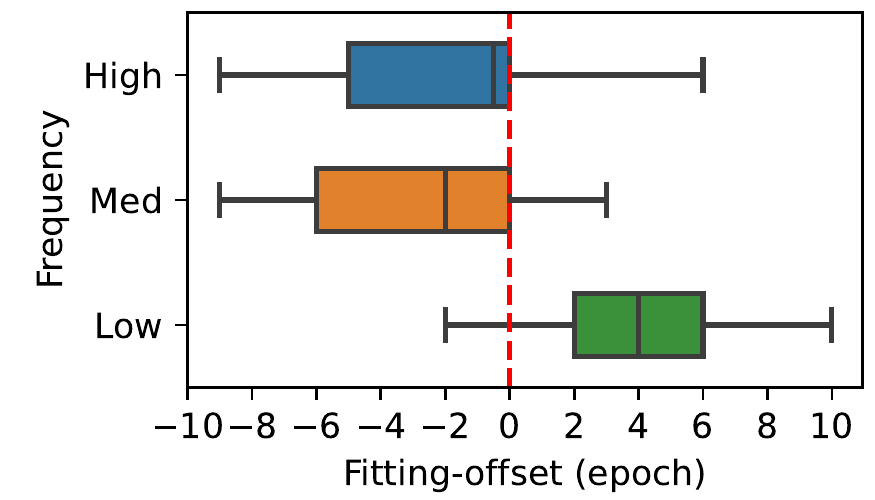}
        \caption{Fitting-offset of tokens grouped by \emph{frequency}.}
        \label{fig:fitdist-freq-de2en}
    \end{subfigure}
    \hfill
    \begin{subfigure}[b]{0.325\linewidth}
        \centering
        \includegraphics[width=1\linewidth]{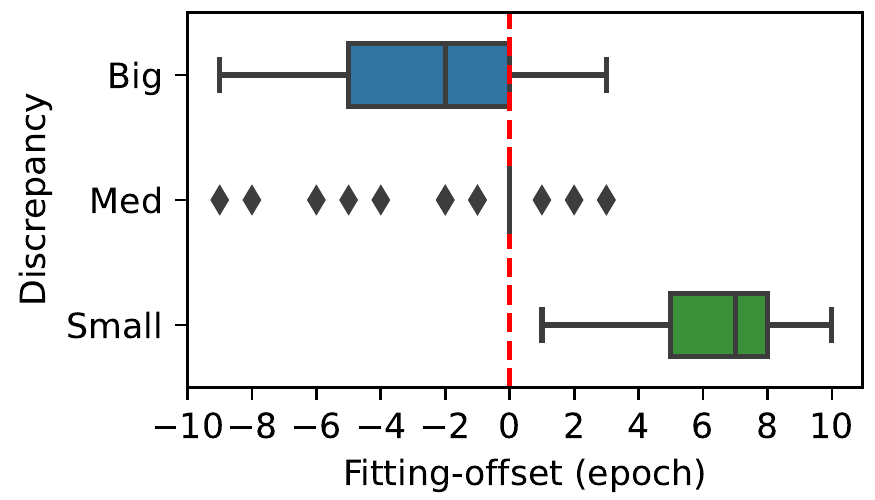}
        \caption{Fitting-offset of tokens grouped by \emph{discrepancy}.}
        \label{fig:fitdist-diff-de2en}
    \end{subfigure}
    \hfill
    \begin{subfigure}[b]{0.325\linewidth}
        \centering
        \includegraphics[width=1\linewidth]{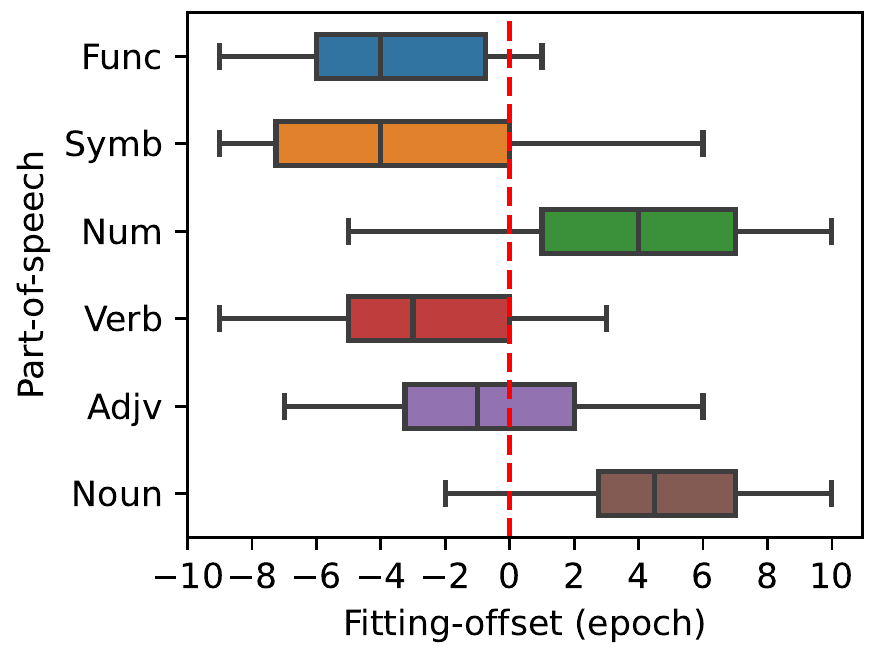}
        \caption{Fitting-offset of tokens grouped by \emph{parts-of-speech}.}
        \label{fig:fitdist-pos-de2en}
    \end{subfigure}
    \caption{The Language: Fitting-offset of English tokens evaluated on News \emph{German-English} dataset.}
    \label{fig:fitdist-de2en}
\end{figure*}

\begin{figure*}[h]
    \begin{minipage}{0.44\linewidth}
        \centering
        \includegraphics[width=1\linewidth]{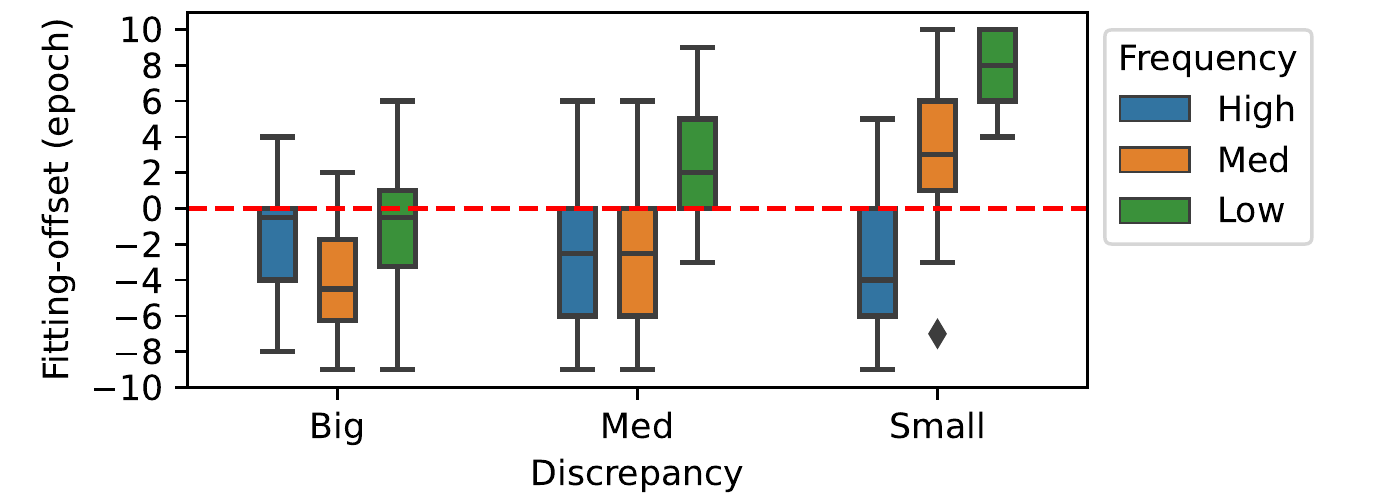}
        \caption{The Language: Fitting-offset of tokens grouped by \emph{discrepancy} and \emph{frequency}.}
        \label{fig:fitdist-freq-diff-de2en}
    \end{minipage}
    \hfill
    \begin{minipage}{0.54\linewidth}
        \centering\small
        \begin{tabular}{@{}lccc@{}}
            \hline
            \multirow{2}{*}{\bf{Discrepancy}} & \multicolumn{3}{c}{\bf{Frequency}}  \\
             & \bf{High} & \bf{Med} & \bf{Low} \\
            \hline
            Big & 67.14 +0.21 & 27.14 -0.27 & 9.90 +0.20 \\
            Med & 69.91 +0.07 & 47.08 -0.08 & 14.67 \bf{+0.78} \\
            Small & 62.99 +0.33 & 81.85 +0.33 & 66.38 \bf{+1.35} \\
            \hline
        \end{tabular}
        \captionof{table}{The Language: Potential-gain for each category grouped by \emph{frequency} and \emph{discrepancy}.}
        \label{tab:fitdist-freq-diff-de2en}
    \end{minipage}
\end{figure*}

\begin{figure*}[h]
    \centering
    \includegraphics[width=1\linewidth]{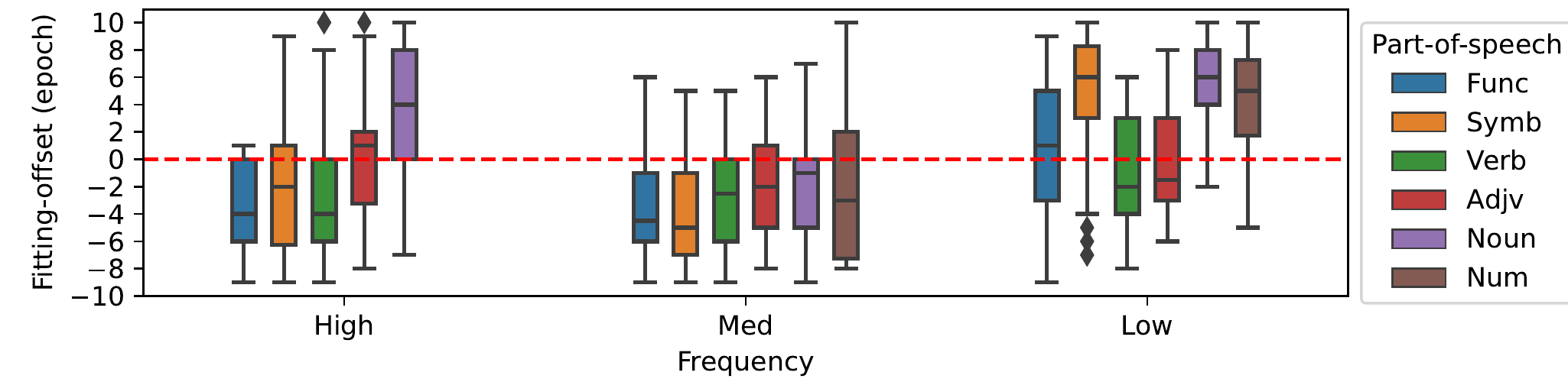}
    \caption{The Language: Fitting-offset of tokens grouped by \emph{frequency} and \emph{POS}.}
    \label{fig:fitdist-freq-pos-de2en}
\end{figure*}

\begin{table*}[h]
    \centering\small
    \begin{tabular}{lcccccc}
        \hline
        \bf{Frequency} & \bf{Function} & \bf{Symbol} & \bf{Number} & \bf{Verb} & \bf{Adj/Adv} & \bf{Noun} \\
        \hline
        High & 64.8 +0.15 & 80.08 \bf{+0.86} & nan +nan & 64.51 \bf{+1.20} & 38.24 \bf{+1.93} & 49.15 \bf{+4.01} \\
        Med & 49.82 +0.13 & 66.23 \bf{+1.67} & 69.43 \bf{+1.29} & 47.03 +0.38 & 55.78 +0.34 & 61.54 +0.21 \\
        Low & 11.12 \bf{+1.75} & 36.09 \bf{+5.98} & 72.28 \bf{+1.29} & 30.92 \bf{+0.57} & 37.76 \bf{+0.78} & 42.13 \bf{+1.20} \\
        \hline
    \end{tabular}
    \caption{The Language: Potential-gain for tokens grouped by \emph{frequency} and \emph{POS}.}
    \label{tab:fitdist-freq-pos-de2en}
\end{table*}

We present the experimental results of the additional factors here, prefixing the name of each factor on the caption.

\begin{figure*}[h]
    \centering
    \includegraphics[width=1\linewidth]{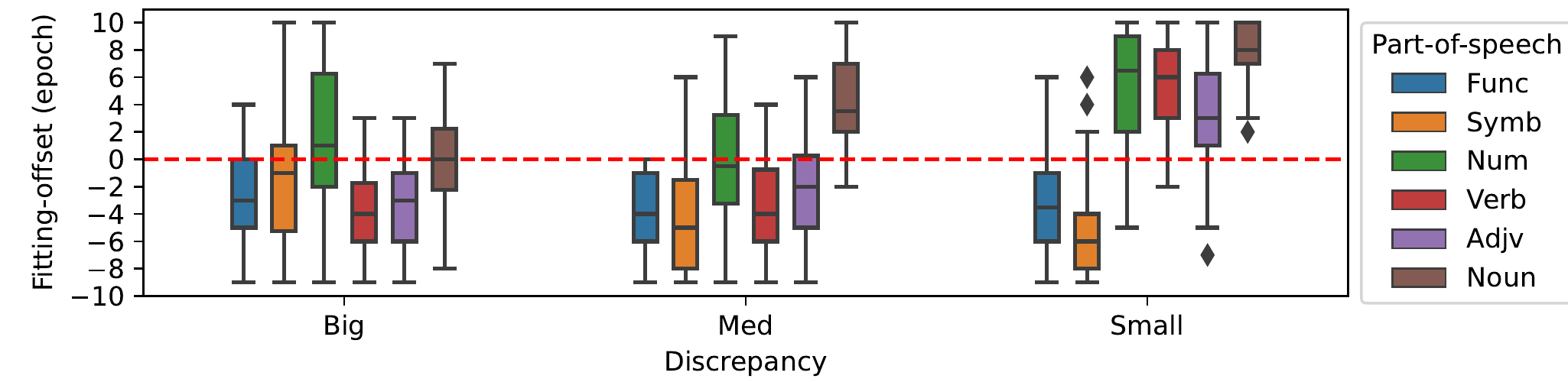}
    \caption{The Language: Fitting-offset of tokens grouped by \emph{discrepancy} and \emph{POS}.}
    \label{fig:fitdist-diff-pos-de2en}
\end{figure*}

\begin{table*}[h]
    \centering\small
    \begin{tabular}{lcccccc}
        \hline
        \bf{Discrepancy} & \bf{Function} & \bf{Symbol} & \bf{Number} & \bf{Verb} & \bf{Adj/Adv} & \bf{Noun} \\
        \hline
        Big & 55.19 +0.10 & 77.67 \bf{+0.58} & 28.34 \bf{+1.78} & 20.76 +0.04 & 15.50 -0.20 & 16.61 +0.19 \\
        Med & 63.84 +0.03 & 75.72 \bf{+1.05} & 38.57 \bf{+2.92} & 32.44 -0.10 & 29.21 +0.35 & 26.49 \bf{+0.92} \\
        Small & 71.99 +0.20 & 75.22 \bf{+1.32} & 86.89 \bf{+1.08} & 65.93 \bf{+1.14} & 72.88 \bf{+0.81} & 72.12 \bf{+1.20} \\
        \hline
    \end{tabular}
    \caption{The Language: Potential-gain for tokens grouped by \emph{discrepancy} and \emph{POS}.}
    \label{tab:fitdist-diff-pos-de2en}
\end{table*}

\begin{figure*}[h]
    \begin{subfigure}[b]{0.325\linewidth}
        \centering
        \includegraphics[width=1\linewidth]{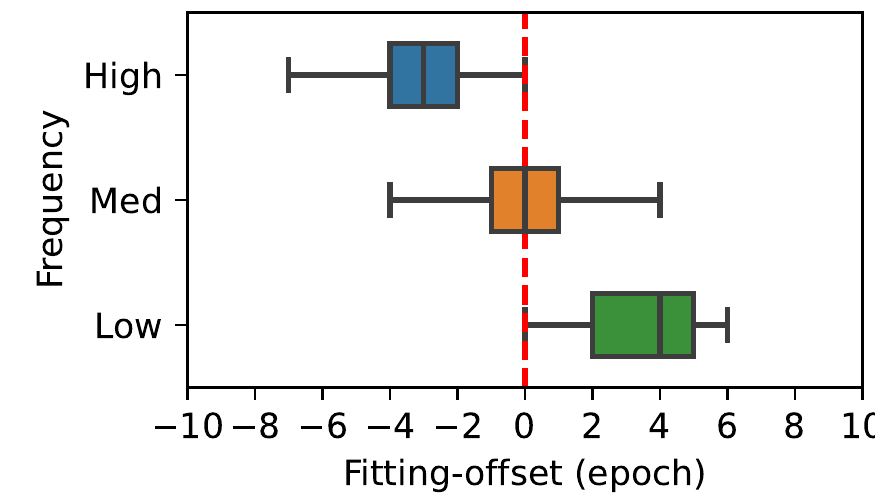}
        \caption{Fitting-offset of tokens grouped by \emph{frequency}.}
        \label{fig:fitdist-freq-en2de-big}
    \end{subfigure}
    \hfill
    \begin{subfigure}[b]{0.325\linewidth}
        \centering
        \includegraphics[width=1\linewidth]{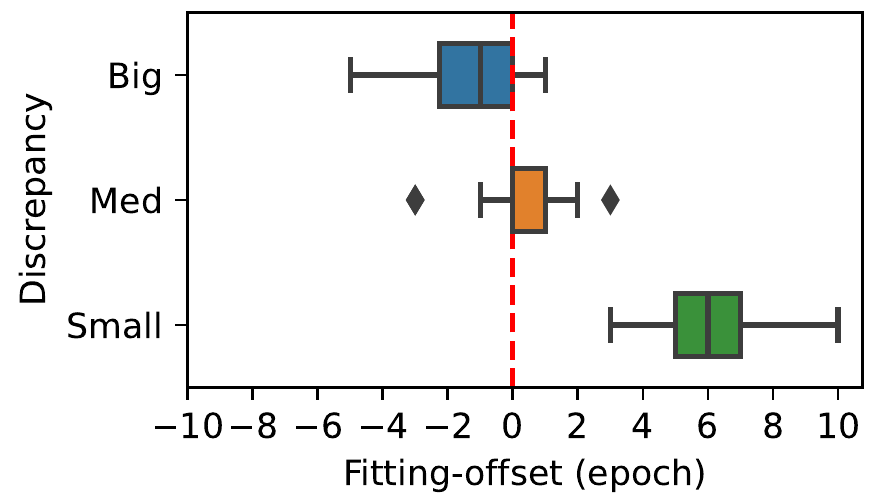}
        \caption{Fitting-offset of tokens grouped by {discrepancy}.}
        \label{fig:fitdist-diff-en2de-big}
    \end{subfigure}
    \hfill
    \begin{subfigure}[b]{0.325\linewidth}
        \centering
        \includegraphics[width=1\linewidth]{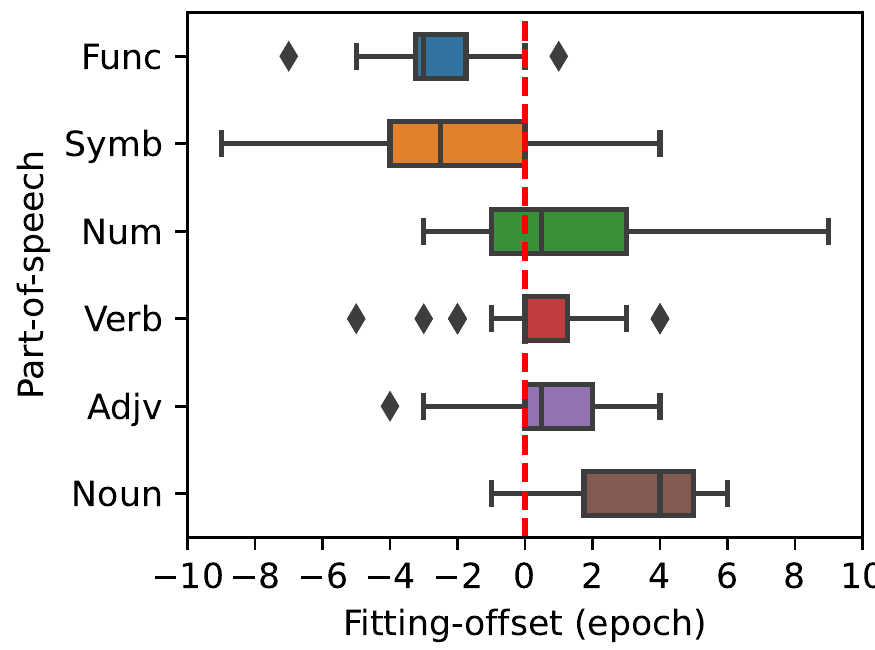}
        \caption{Fitting-offset of tokens grouped by \emph{parts-of-speech}.}
        \label{fig:fitdist-pos-en2de-big}
    \end{subfigure}
    \caption{The Model Size: Fitting-offset of German tokens evaluated on News English-German with \emph{big} model.}
    \label{fig:fitdist-en2de-big}
\end{figure*}

\newpage

\begin{figure*}[h]
    \begin{subfigure}[b]{0.325\linewidth}
        \centering
        \includegraphics[width=1\linewidth]{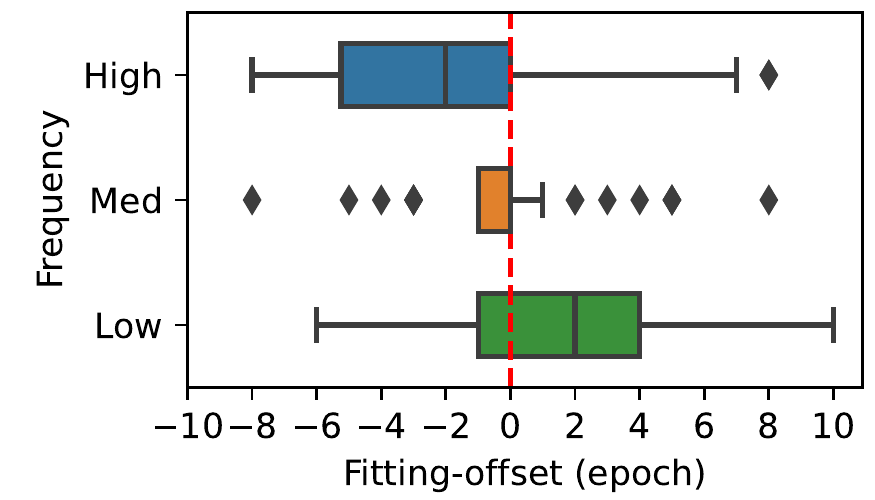}
        \caption{Fitting-offset of tokens grouped by \emph{frequency}.}
        \label{fig:fitdist-freq-en2de-europarl25w}
    \end{subfigure}
    \hfill
    \begin{subfigure}[b]{0.325\linewidth}
        \centering
        \includegraphics[width=1\linewidth]{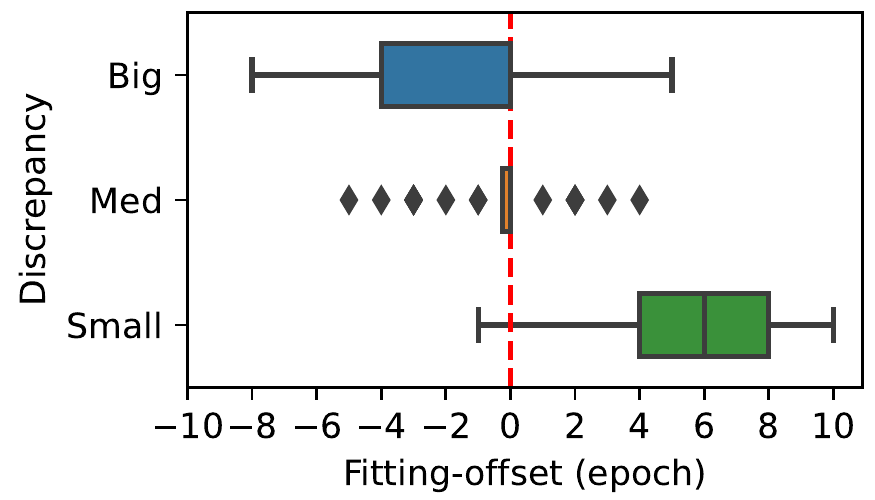}
        \caption{Fitting-offset of tokens grouped by \emph{discrepancy}.}
        \label{fig:fitdist-diff-en2de-europarl25w}
    \end{subfigure}
    \hfill
    \begin{subfigure}[b]{0.325\linewidth}
        \centering
        \includegraphics[width=1\linewidth]{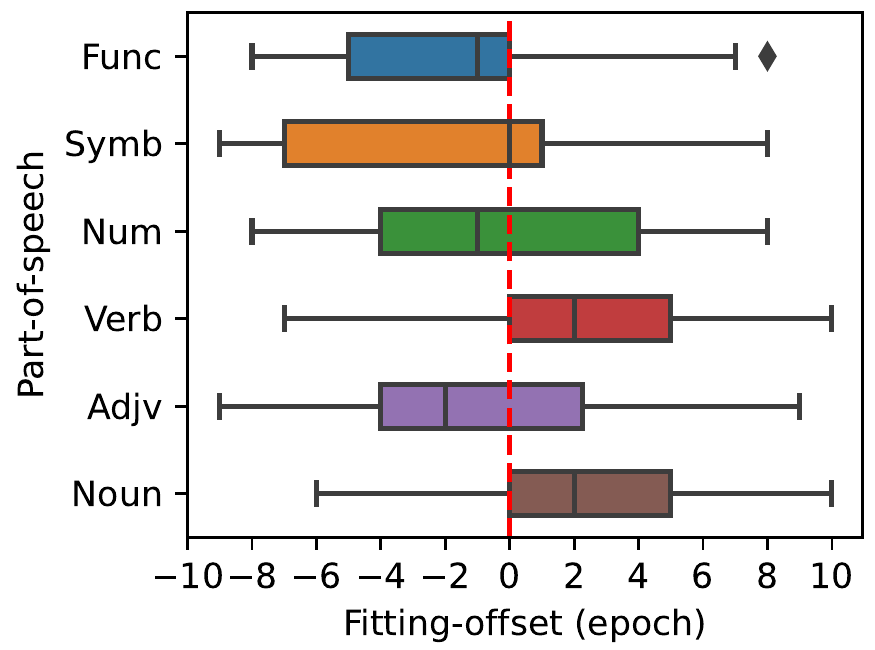}
        \caption{Fitting-offset of tokens grouped by \emph{parts-of-speech}.}
        \label{fig:fitdist-pos-en2de-europarl25w}
    \end{subfigure}
    \caption{The Domain: Fitting-offset of tokens evaluated on \emph{Europarl} English-German ($250,000$ samples).}
    \label{fig:fitdist-en2de-europarl25w}
\end{figure*}

\begin{figure*}[h]
    \begin{subfigure}[b]{0.325\linewidth}
        \centering
        \includegraphics[width=1\linewidth]{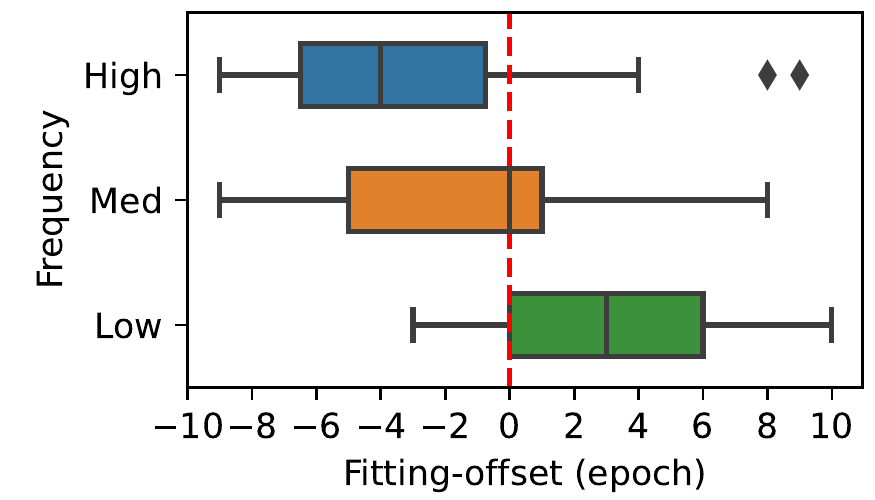}
        \caption{Fitting-offset of tokens grouped by \emph{frequency}.}
        \label{fig:fitdist-freq-en2de-europarl50w}
    \end{subfigure}
    \hfill
    \begin{subfigure}[b]{0.325\linewidth}
        \centering
        \includegraphics[width=1\linewidth]{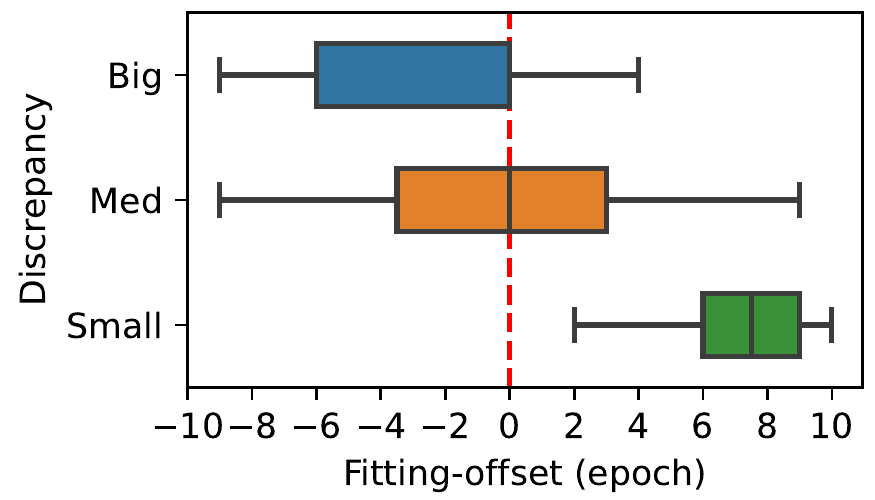}
        \caption{Fitting-offset of tokens grouped by \emph{discrepancy}.}
        \label{fig:fitdist-diff-en2de-europarl50w}
    \end{subfigure}
    \hfill
    \begin{subfigure}[b]{0.325\linewidth}
        \centering
        \includegraphics[width=1\linewidth]{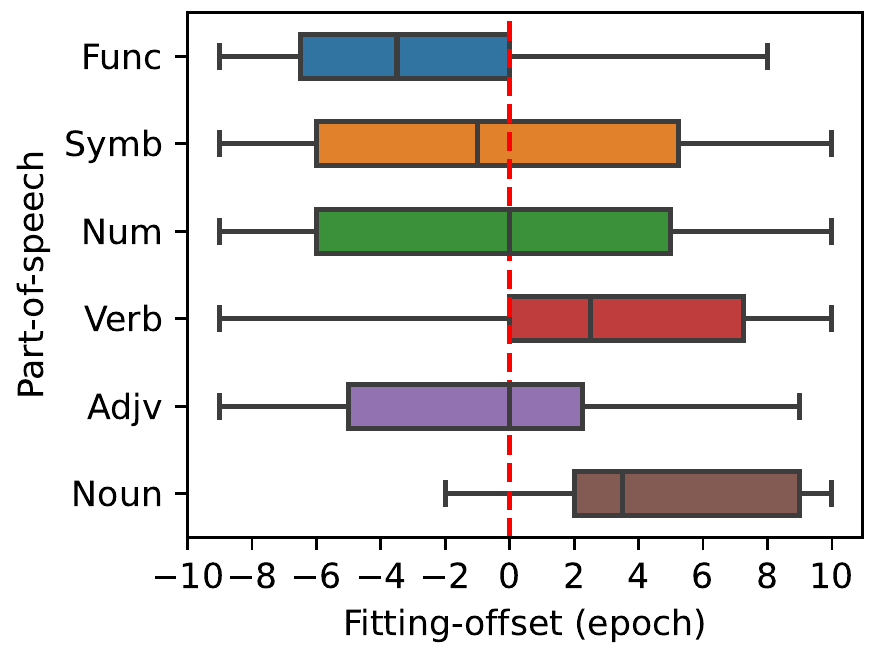}
        \caption{Fitting-offset of tokens grouped by \emph{parts-of-speech}.}
        \label{fig:fitdist-pos-en2de-europarl50w}
    \end{subfigure}
    \caption{The Data Scale: Fitting-offset of tokens evaluated on Europarl English-German (\emph{$500,000$ samples})..}
    \label{fig:fitdist-en2de-europarl50w}
\end{figure*}

\end{document}